# Action Recognition in Untrimmed Videos with Composite Self-Attention Two-Stream Framework

Dong Cao[1,2], Lisha Xu[1,2], and HaiBo Chen[2]

[1] Institute of Cognitive Intelligence, DeepBlue Academy of Sciences
doocao@gmail.com, xuls@deepblueai.com
[2] DeepBlue Technology (Shanghai) Co., Ltd. No.369, Weining Road, Shanghai, China
chenhaibo@deepblueai.com

**Abstract.** With the rapid development of deep learning algorithms, action recognition in video has achieved many important research results. One issue in action recognition, Zero-Shot Action Recognition (ZSAR), has recently attracted considerable attention, which classify new categories without any positive examples. Another difficulty in action recognition is that untrimmed data may seriously affect model performance. We propose a composite two-stream framework with a pre-trained model. Our proposed framework includes a classifier branch and a composite feature branch. The graph network model is adopted in each of the two branches, which effectively improves the feature extraction and reasoning ability of the framework. In the composite feature branch, a 3-channel self-attention models are constructed to weight each frame in the video and give more attention to the key frames. Each self-attention models channel outputs a set of attention weights to focus on a particular aspect of the video, and a set of attention weights corresponds to a one-dimensional vector. The 3-channel self-attention models can evaluate key frames from multiple aspects, and the output sets of attention weight vectors form an attention matrix, which effectively enhances the attention of key frames with strong correlation of action. This model can implement action recognition under zero-shot conditions, and has good recognition performance for untrimmed video data. Experimental results on relevant data sets confirm the validity of our model.

## 1 Introduction

### 1.1 Action recognition

In computer vision, video understanding is one of the most important fields. As a brunch of video understanding, action recognition is the base of visual reasoning tasks, e.g. video captioning, video relational reasoning. Current mainstream models for action recognition are almost derived from Two-stream Convolutional Network and Convolution-3D Network.

Some studies have found that inputting optical flow into 2D CNNs made a better performance than merely RGB frames as optical flow represent the motion information at some extent. Thus it is grant to propose Two-stream networks, which take

RGB images and optical flow as input, providing the state of the art result[1, 11]. One of the derivation of Two-stream networks, Temporal Segment Network(TSN), uses multiple two-stream modules to learn parameters effectively and efficiently with combining video-level supervision and sparse temporal sampling strategy, even given limited training samples, resulting in good performance in long-term temporal structure[18].

However, since optical flow is computation-expensive and supposed to calculate offline, it cannot meet the demand of real-time application. 3D spatiotemporal CNN models which use 3D ConvNet enable model to extract spatial features and temporal features simultaneously, over 10 times faster than two-stream even though sacrifice some accuracy, have drawn more attention[8].

Basic 3D spatiotemporal CNNs suffer from large parameters and are not easy to train from scratch. Therefore, a model named I3D, based on the pre-trained model on ImageNet, inflates the 2D ConvNets into 3D to increase the temporal dimension, having much fewer parameters. Furthermore, I3D found that two-stream design still improve the accuracy[1]. Another (2+1)D architecture decomposes 3D convolutional filters into separate spatial and temporal components, so that renders the optimization, maintaining the solid performance with 2D CNNs in single frame and speeding up the calculation[15].

To balance the accuracy and speed, some works also adjust the input data of two stream. Like extracting key-frames for spatial stream[20], or processing kinds of optical flow information for temporal stream[2, 3, 13], relax the burden of calculation and store.

### 1.2 Zero-Shot learning

Zero-shot learning aims to construct new classifiers dynamically based on the semantic descriptors provided by human beings or the existing knowledge, rather than labeled data. Due to poor scalability of exhaustive annotation for supervised learning [19] and expectation to emulate the human ability to learn semantics with few examples, zero-shot have drawn considerable attention. There have been various architectures proposed for zero-shot action recognition (ZSAR). In these studies, the information of unseen classes come from human-defined attribution, language in the form of text descriptions or word embedding, effective in trimmed videos datasets. In this work, we propose a novel architecture that perform well in untrimmed videos.

## 2 Related Work

**Zero-shot action recognition** With the explosion of videos and the difficulty of scalability of manual annotation, there have been approaches to recognize the unknown action in the way of zero-shot. ZSAR associates known categories and unknown categories based on semantic space, including manual attributes [23, 24, 25, 26, 27], text descriptions [28] and word vectors [29, 30]. However, attributes are not easy to define and the manually-specified attributes are highly subjective. Thus, it is hard to generate ambiguous and undefined categories in real scenes. Word embeddings are more objective for addressing ZSAR. [19, 39] embedded the labels of videos as word vectors for a shared semantic space. [31] utilizes both category-level semantics and intrinsic data structures by adopting error-correcting output codes to address domain shift problem. But these methods usually ignore the temporal information of videos, which take significant advantages for visual understanding[32]. [5] proposed a zero-shot action recognition framework using both the visual clues and external knowledge to show relations between objects and actions, also applied self-attention to model the temporal information of videos.

**Transfer learning** In many real-world applications, it is expensive to re-collect training data and re-model when task changes [33]. Knowledge transfer or transfer learning is needed between task domains, research referring has drawn increasing attention.[34] using the similarity of classifier in source and target Domain to calculate the gradient between them. [35, 36] used a max mean discrepancies (MMD) as the distance between source and target. Recently, the study about domain adaption are expected to improve the knowledge transfer [37].

# 3 Our Approach

In this section, we present the details of our proposed model. The method is inspired by literatures[5, 38]. The novel model consists of a composite two-stream structure, and including recognition branch and composite feature branch. A sub-dual-branch structure is constructed in the composite feature branch. Each of the sub-dual-branch structures adopts a multi-channel self-attention model for feature extraction and action recognition of untrimmed and trimmed video respectively.

We introduce the novel model from three levels as follows. In section 3.1, the first level is the sub-branch of multi-channel self-attention model for trimmed video. In section 3.2, the second level is the model of composite feature branch used for both trimmed and untrimmed video. In section 3.3, the third level is the composite two-stream structure for zero-shot action recognition, whether it is trimming video or untrimmed video data sets.

## 3.1 Sub-Branch of Multi-Channel Self-Attention Model

**Video data preprocessing**

The model processes trimmed video data, and implements action recognition and classification of the trimmed video data. Firstly, the original video data needs to be preprocessed. The purpose of the preprocessing is to extract the features of the video, including two aspects. The first is feature extraction from the spatial dimension, and the second is feature extraction in the temporal dimension.

Specifically, the feature extraction in the first aspect is the extraction of RGB information in the spatial dimension. For each frame of the original video having RGB three-channel information, respectively, through a deep neural network model, the deep neural network can select the ResNet101 network, where we use the pre-trained ResNet101 network to directly perform feature extraction tasks. The output is an s-dimensional vector corresponding to each frame. Giving a video data contains G frames, G s-dimensional vectors are output, and a $s \times G$ matrix is formed sequentially. The matrix denoted as $\mathcal{S} \in \mathbb{R}^{s \times G}$, we named this first frame as a spatial feature extractor.

The second aspect of feature extraction is the extraction of optical flow information in the temporal dimension. First, the original video data is processed by the general optical flow algorithm, and the optical flow data is input to another independent deep neural network model. Here we adopt pre-trained ResNet101 network, where the ResNet101 network is independent of the network in the first extractor above. After the feature extraction of the ResNet101 network, the output is a t-dimensional vector corresponding to each optical stream data. Then the same piece of video data is included G frames, and (G-1) t-dimensional vectors can be obtained, which form a matrix of $t \times (G - 1)$, denoted as $\mathcal{T} \in \mathbb{R}^{t \times (G-1)}$. We name this second frame as a temporal feature extractor.

**Multi-channel self-attention model**

The result of data preprocessing consists of two parts. The first part is the spatial feature information matrix $\mathcal{S}$ obtained by extracting the original video from the spatial feature extractor. The second part is the temporal feature information matrix $\mathcal{T}$ obtained from the temporal feature extractor and the original video. Then, the spatial feature information matrix $\mathcal{S}$ and the temporal feature information matrix $\mathcal{T}$ are respectively input two independent multi-channel self-attention models, and the output result is two information matrix weighted by attention, corresponding to the input spatial feature information matrix $\mathcal{S}$ and temporal feature information matrix $\mathcal{T}$. These two matrixes are denoted as $\mathbb{S}$ and $\mathbb{T}$, respectively[38]. The multi-channel self-attention model consists of two independent parts, which are respectively applied to the spatial feature information matrix $\mathcal{S}$ and the temporal feature information matrix $\mathcal{T}$, named as multi-channel spatial self-attention (MC-spatial Self-Attention) and multi-channel temporal self-attention (MC-temporal Self-Attention).

The structure of the MC-spatial Self-Attention is shown in the left part of Figure 1. The 3-channel self-attention structure is used. The spatial self-attention structure of each channel is the same, but different activation functions are used, for example, sigmod, tanh, and Leaky ReLU can be selected to facilitate the extraction of spatial information from different angles of attention to form a multi-angle attention weight vector. The data input passes through the fully connected layer FC11, the activation function, the fully connected layer FC12 and softmax. Fully connected layer FC11, corresponding to parameters $W_{1j} \in \mathbb{R}^{a \times s}, j = 1{,}2{,}3$, $a$ is a configurable hyperparameter; activation functions are sigmod, tanh, Leaky ReLU; Fully connected layer FC12, corresponding the parameter $u_{1j} \in \mathbb{R}^{1 \times a}, j = 1{,}2{,}3$; The softmax output results in the attention vector $\mathrm{V}_{1j}$, where $\mathrm{V}_{1j} \in \mathbb{R}^{G \times 1}, j = 1{,}2{,}3$. The operation relationship is as follows.

$$\mathrm{V}_{11} = softmax[u_{11} \cdot sigmoid(W_{11}\mathcal{S})]$$
$$\mathrm{V}_{12} = softmax[u_{12} \cdot tanh(W_{12}\mathcal{S})]$$
$$\mathrm{V}_{13} = softmax[u_{13} \cdot LeakyReLU(W_{13}\mathcal{S})]$$

Further $\mathrm{V}_{1j} \in \mathbb{R}^{G\times 1}, j = 1,2,3$ are multiplied by the matrix of the spatial feature matrix $\mathcal{S}$ respectively, to obtain the attentional space feature matrix $\mathbb{S}_i$, where $\mathbb{S}_i \in \mathbb{R}^{s\times G}, i = 1,2,3$.

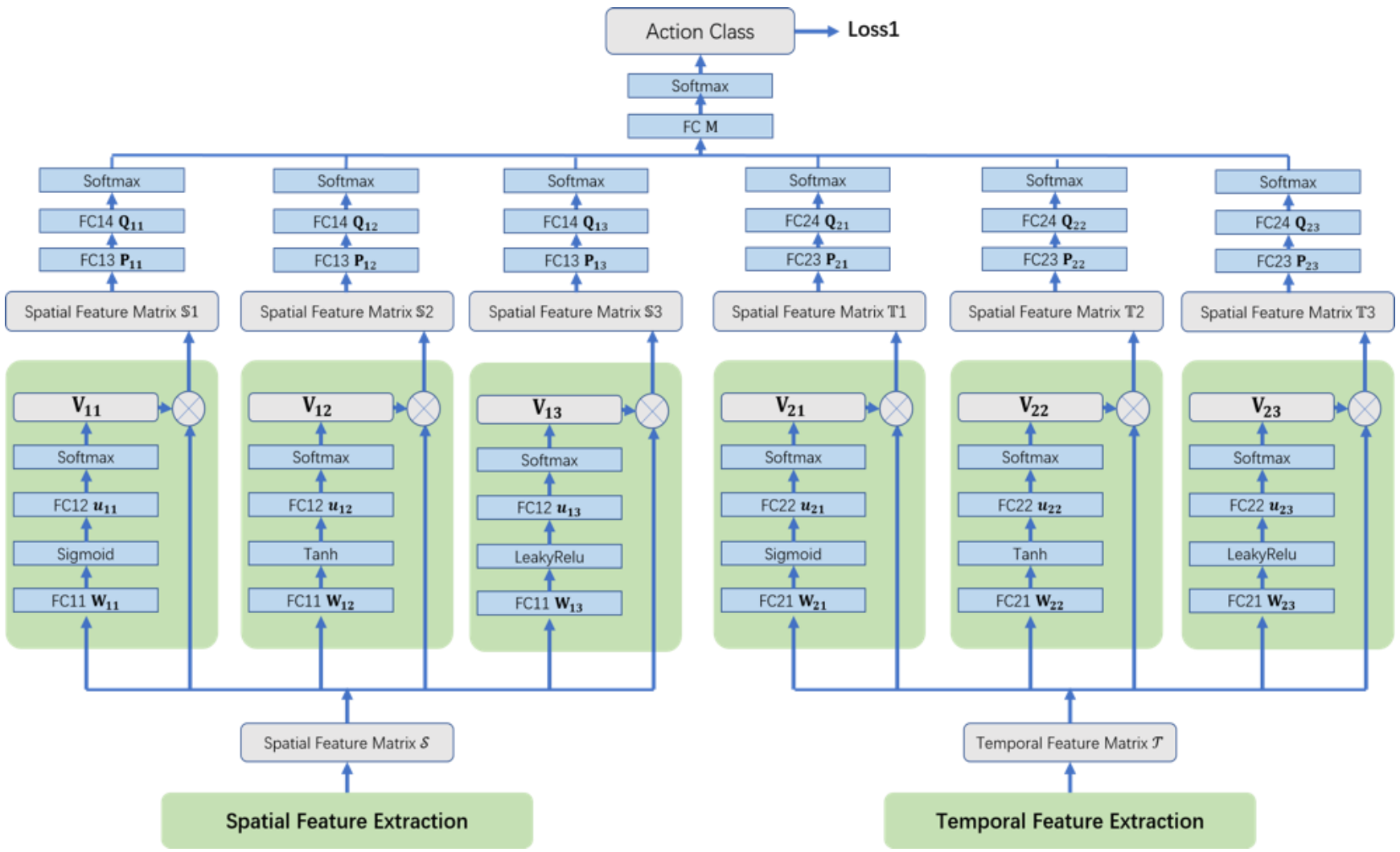

**Fig. 1.** The structure of multi-channel self-attention models

The structure of the MC-temporal Self-Attention is shown in the right part of Figure 1. The 3-channel self-attention structure is also used. The temporal self-attention structure of each channel is the same. Different activation functions are used. In order to facilitate the extraction of temporal information from different angles of attention, forming a multi-angle attention weight vector. The data input passes through the fully connected layer FC21, the activation function, the fully connected of FC22 and softmax. The fully connected layer FC21, corresponding parameters $W_{2j} \in \mathbb{R}^{b\times t}, j = 1,2,3$, $b$ is configurable hyperparameter; Activation functions are sigmod, tanh, Leaky ReLU; The fully connected FC22, corresponding parameters $u_{2j} \in \mathbb{R}^{1\times b}, j = 1,2,3$; the softmax output results in the attention vector $\mathrm{V}_{2j}$, where$\mathrm{V}_{2j} \in \mathbb{R}^{(G-1)\times 1}, j = 1,2,3$. The operation relationship is as follows.

$$\mathrm{V}_{21} = softmax[u_{21} \cdot sigmoid(W_{21}\mathcal{T})]$$
$$\mathrm{V}_{22} = softmax[u_{22} \cdot tanh(W_{22}\mathcal{T})]$$
$$\mathrm{V}_{23} = softmax[u_{23} \cdot LeakyReLU(W_{23}\mathcal{T})]$$

Further $\mathrm{V}_{2j} \in \mathbb{R}^{(G-1)\times 1}, j = 1,2,3$ are multiplied by the temporal feature matrix $\mathcal{T}$, respectively, to obtain the attentional temporal feature matrix $\mathbb{T}_i$, where $\mathbb{T}_i \in \mathbb{R}^{t\times (G-1)}, i = 1,2,3$.

For the attention spatial feature matrix $\mathbb{S}_i$ and the attention temporal feature matrix $\mathbb{T}_i$ obtained by the above operation, the following operations are performed, which are used to calculate the loss function in the training phase, expressed as $loss1$, and used for action recognition in the inference phase.

$$loss1, inference \Leftrightarrow softmax\{M\{[softmax[Q_{1i}(P_{1i}\mathbb{S}_i)^T]] \parallel [softmax[Q_{2i}(P_{2i}\mathbb{T}_i)^T]]\}^T\}$$

Where $P_{1i} \in \mathbb{R}^{1\times s}, Q_{1i} \in \mathbb{R}^{k\times G}, i = 1,2,3$, $P_{2i} \in \mathbb{R}^{1\times t}, Q_{2i} \in \mathbb{R}^{k\times(G-1)}, M \in \mathbb{R}^{1\times 6}$. The symbol $\parallel$ operation means that multiple column vectors are sequentially combined side by side into a matrix, where three column vectors of spatial features and three column vectors of temporal features are combined side by side to obtain $k \times 6$ matrix. $k$ is the number of classifications for action recognition.

### 3.2 Model of Composite Feature Branch

The multi-channel self-attention model in the previous section is used to handle the action recognition problem of trimming video. However, in the face of untrimmed video, action recognition performance of the model is significantly affected by the inclusion of several background frames in such videos. In order to solve this problem, we construct a composite feature branch model[38], as shown in Figure 2 below.

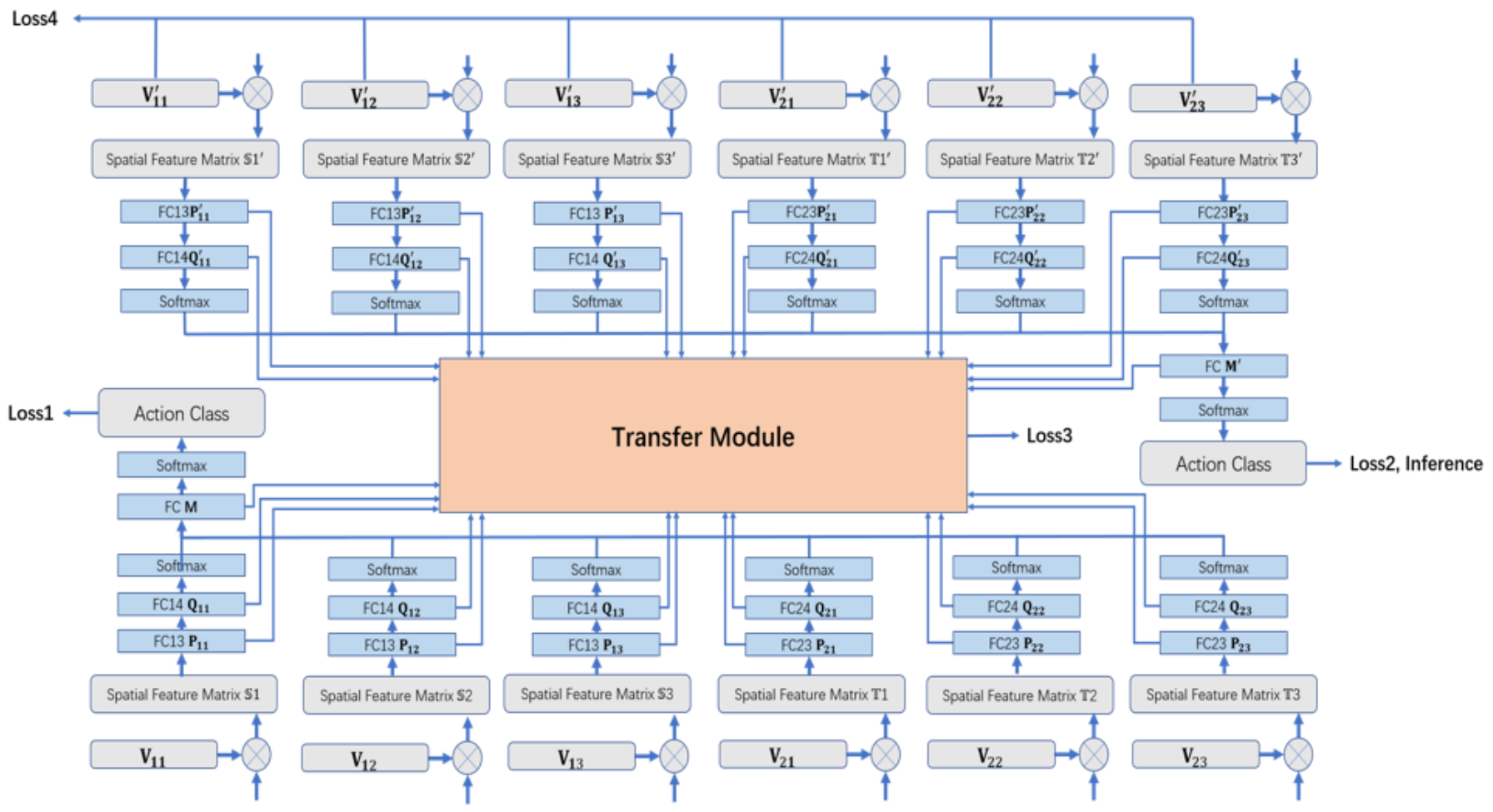

**Fig. 2.** Composite feature branch model

We designed the two-stream parallel composite architecture with multi-channel self-attention model. The upper stream is used to process untrimmed videos, and the lower stream is used to process trimmed videos.

The upper part deals with the model of untrimmed videos, where the loss function defined in the output part of the action classification is expressed as loss2, which is calculated using the form of standard multi-class cross entropy, which is related to the following representation.

$$loss2, inference \Leftrightarrow softmax\{M'\{[softmax[Q'_{1i}(P'_{1i}\mathbb{S}'_i)^T]] \parallel [softmax[Q'_{2i}(P'_{2i}\mathbb{T}'_i)^T]]\}^T\}$$

Where $P'_{1i} \in \mathbb{R}^{1\times s}, Q'_{1i} \in \mathbb{R}^{k\times G}, i = 1,2,3,\ P'_{2i} \in \mathbb{R}^{1\times t}, Q'_{2i} \in \mathbb{R}^{k\times (G-1)}, M' \in \mathbb{R}^{1\times 6}$, here are the three-column vectors of temporal features and spatial features. The three column vectors are combined side by side to obtain a $k \times 6$ matrix. $k$ is the number of classifications for action recognition.

Here's how to use the knowledge of the trimmed videos dataset to improve the performance of the model on the untrimmed videos dataset through transfer learning.

First of all, the model learning on the trimmed videos data set, specifically, is to train the lower stream model of Figure 2, the detailed structure of the lower stream is shown in Figure 2, the loss function is loss1, the specific form is standard classification cross entropy form, see expression "loss1, inference". After the training of the lower stream model is completed, the parameters in the model learn the quasi-optimal distribution values, which can be used to infer the data of the trimmed videos. In this part, with the multi-channel spatial self-attention model and the multi-channel temporal self-attention model, the classifier related parameters in a total of six channels are transmitted to the transfer module, including: P11, P12, P13, in the fully connected layer FC13; Q11, Q12, Q13 in the fully connected layer FC14; P23, P22, P23 in the fully connected layer FC23, Q21, Q22, Q23 in the fully connected layer FC24. M in the fully connected layer FC.

Then, with the multi-channel spatial self-attention model and the multi-channel temporal self-attention model of the upper stream, the corresponding classifier parameters of the total of 6 channels are transmitted to the transfer module in real time, including: $P'_{11}, P'_{12}, P'_{13}$ in the fully connected layer FC13, $Q'_{11}, Q'_{12}, Q'_{13}$ in the fully connected layer FC14, $P'_{21}, P'_{22}, P'_{23}$ in the fully connected layer FC23, $Q'_{21}, Q'_{22}, Q'_{23}$ in the fully connected layer FC24, $M'$ in the fully connected layer FC. The loss function loss3 of the transfer module is constructed by calculating the generalized maximum mean discrepancy (MMD) of the upper and lower models by the literature [35].

In the third step, during the training process of the upper stream, the attention vectors $V'_{ij}, i = 1,2, j = 1,2,3,$ need to be regularization to optimize the structure of the loss function loss4.

Given $V'_{1j} \in \mathbb{R}^{G\times 1}, V'_{2j} \in \mathbb{R}^{(G-1)\times 1}, j = 1,2,3$ , let $V'_{1j} = (v'_{1j1}, v'_{1j2}, \cdots, v'_{1jG})^T$ , $V'_{2j} = (v'_{2j1}, v'_{2j2}, \cdots, v'_{2j(G-1)})^T$, where T represents a transposition operation. Then $loss4$ is calculated as follows:

$$loss4 = \sum_{j=1}^{3}\sum_{n=1}^{G-1}(v'_{1jn} - v'_{1j(n+1)})^4 + \sum_{j=1}^{3}\sum_{n=1}^{G-2}(v'_{2jn} - v'_{2j(n+1)})^4 + \sum_{j=1}^{3}(\parallel V'_{1j} \parallel_1 + \parallel V'_{2j} \parallel_1)$$

$$= \sum_{j=1}^{3}\{2((V'_{1j})^T \circ (V'_{1j})^T)(V'_{1j} \circ V'_{1j}) - ((V'_{1j})^T \circ (V'_{1j})^T)\mathcal{A}(V'_{1j} \circ V'_{1j}) - ((V'_{1j})^T \circ (V'_{1j})^T)\mathcal{B}(V'_{1j} \circ V'_{1j}) + 6((V'_{1j})^T \circ (V'_{1j})^T)\mathcal{C}(V'_{1j} \circ V'_{1j}) - 4((V'_{1j})^T \circ (V'_{1j})^T \circ (V'_{1j})^T)\mathcal{C}V'_{1j} - 4(V'_{1j})^T((\mathcal{C}V'_{1j}) \circ (\mathcal{C}V'_{1j}) \circ (\mathcal{C}V'_{1j}))\}$$

$$
\begin{aligned}
&+\sum_{j=1}^{3}\{2((V'_{2j})^T \circ (V'_{2j})^T)(V'_{2j} \circ V'_{2j}) - ((V'_{2j})^T \circ (V'_{2j})^T)\mathcal{D}(V'_{2j} \circ V'_{2j}) - ((V'_{2j})^T \\
&\quad \circ (V'_{2j})^T)\mathcal{E}(V'_{2j} \circ V'_{2j}) + 6((V'_{2j})^T \circ (V'_{2j})^T)\mathcal{F}(V'_{2j} \circ V'_{2j}) - 4((V'_{2j})^T \\
&\quad \circ (V'_{2j})^T \circ (V'_{2j})^T)\mathcal{F}V'_{2j} - 4(V'_{2j})^T((\mathcal{F}V'_{2j}) \circ (\mathcal{F}V'_{2j}) \circ (\mathcal{F}V'_{2j}))\} \\
&+\sum_{j=1}^{3}(\| V'_{1j} \|_1 + \| V'_{2j} \|_1)
\end{aligned}
$$

Where $\circ$ represents hadamard product, $\| \|_1$ represents 1-norm, matrices $\mathcal{A}, \mathcal{B}, \mathcal{C}, \mathcal{D}, \mathcal{E}, \mathcal{F}$ are as follows:

$$
\mathcal{A} = \begin{bmatrix} 1 & 0 & & & \\ 0 & 0 & & & \\ & & \ddots & & \\ & & & 0 & 0 \\ & & & 0 & 0 \end{bmatrix}_{G\times G}, \quad
\mathcal{B} = \begin{bmatrix} 0 & 0 & & & \\ 0 & 0 & & & \\ & & \ddots & & \\ & & & 0 & 0 \\ & & & 0 & 1 \end{bmatrix}_{G\times G}, \quad
\mathcal{C} = \begin{bmatrix} 0 & 1 & & & \\ 0 & 0 & & & \\ & & \ddots & & \\ & & & 0 & 1 \\ & & & 0 & 0 \end{bmatrix}_{G\times G},
$$

$$
\mathcal{D} = \begin{bmatrix} 1 & 0 & & & \\ 0 & 0 & & & \\ & & \ddots & & \\ & & & 0 & 0 \\ & & & 0 & 0 \end{bmatrix}_{(G-1)\times(G-1)}, \quad
\mathcal{E} = \begin{bmatrix} 0 & 0 & & & \\ 0 & 0 & & & \\ & & \ddots & & \\ & & & 0 & 0 \\ & & & 0 & 1 \end{bmatrix}_{(G-1)\times(G-1)}, \quad
\mathcal{F} = \begin{bmatrix} 0 & 1 & & & \\ 0 & 0 & & & \\ & & \ddots & & \\ & & & 0 & 1 \\ & & & 0 & 0 \end{bmatrix}_{(G-1)\times(G-1)}
$$

Finally, the resulting loss function for training on untrimmed video data are used for model training on the upper stream, which is set as follows:

$$loss = loss2 + loss3 + loss4$$

### 3.3 Model of Composite Feature Branch

We use the multi-channels self-attention composite model in Section 3.2 as a pre-training model, and then replace the attention module of Two-Stream Graph Convolutional Network in the [5], to implement a multi-channel self-attention composite double-stream model. This model can achieve zero-shot action recognition.

## 4 Experiment

In this section, we evaluate the performance of our novel architecture for zero-shot untrimmed action recognition on THOUMOS14, which contains a large number of trimmed and untrimmed videos.

### 4.1 Experimental setup

**Dataset** The training dataset of THUMOS14 is UCF101, including 101-class trimmed videos. The validation and test dataset of THUMOS14 contain 2500 and 1574 untrimmed videos respectively. Furthermore, the validation is the same as the video category in the training set, while the test data contains only part of 101 categories.

Following the transductive and generalized configuration, we implement our experiments. The former is to access the unseen data in training, while the latter is to combine part of the seen data and the unseen data as test data [12, 39]. Here we design three data splits (TD, G and TD+G), to evaluate the performance of zero-shot action recognition in untrimmed videos.

Since most zero-shot methods currently evaluate their performance on UCF101, we also follow the data split proposed by (Xu, Hospedales, and Gong 2017), i.e. to evaluate our model.

Among the data splitting ways, the setup of TD+G is shown in Table 3. During training, on one hand, we set the 30-class trimmed videos as part of the seen data, responsible for explicit knowledge, which is the source of transfer module. On the other hand, for THUMOS14, we select 80% of other 20-class untrimmed videos as seen data and 20% of remaining 51-class untrimmed videos as unseen data for the target of transfer module. During testing, for THUMOS14, we combine the 20% of those 20-class untrimmed videos and 80% of those 51-class untrimmed videos.

**Table 1.** The split way to evaluate our model in transductive setting.

| THUMOS14 | | | | |
|---|---|---|---|---|
| **TD** | Train | | Test | |
| | Seen | Unseen | Seen | Unseen |
| Trimmed | 30 classes | 0 | 0 | 0 |
| Untrimmed | 20 classes | 51 classes *20% | 0 | 51 classes *80% |

**Table 2.** The split way to evaluate our model in generalized setting

| THUMOS14 | | | | |
|---|---|---|---|---|
| **G** | Train | | Test | |
| | Seen | Unseen | Seen | Unseen |
| Trimmed | 30 classes | 0 | 0 | 0 |
| Untrimmed | 20 classes*80% | 0 | 20 classes*20% | 51 classes |

**Table 3.** The split way to evaluate our model in both transductive and generalized setting.

| THUMOS14 | | | | |
|---|---|---|---|---|
| **TD+G** | Train | | Test | |
| | Seen | Unseen | Seen | Unseen |
| Trimmed | 30 classes | 0 | 0 | 0 |
| Untrimmed | 20 classes*80% | 51 classes *20% | 20 classes*20% | 51 classes *80% |

### 4.2 Evaluation

We evaluate the model in three phases below, and adopt classification accuracy as metric.

1. The trimmed action recognition (TAR) task is to predict the label of trimmed videos. Here we use the training data of THUMOS, and 3 splits of UCF101 to examine the performance of MC-spatial Self-Attention module.

2. The untrimmed action recognition (UTAR) task is to predict the primary action label of untrimmed videos. This task examines the performance of transfer module that learns the explicit knowledge from trimmed videos.
3. The zero-shot action recognition (ZSAR) is to predict the primary action label of untrimmed videos, the categories of which are absent during training.

### 4.3 Implement details

The novel model consists of a composite two-stream structure, and including recognition branch and composite feature branch. For the composite feature branch, we process the optical information by generic end-to-end way, and extract the spatial and temporal features of the input RGB and optical flow information respectively by using the residual network pre-trained in kinetics. In order to reduce the influence caused by different inputs, we uniformly limit the input size to $16 \times 224 \times 224$.

For the recognition branch, we encode action categories through Glove, and construct knowledge graph by ConceptNet. To train the whole model, we optimize the parameter with SGD where initial learning rate is 0.0001 and weight decay 0.0005 every 1000 iterations.

### 4.4 Result

In TAR, we compare our sub-model with some baselines: (1) Two-stream CNN (K. Simonyan et al. 2014) (2) Two-stream+LSTM (J. Y.-H. Ng et al, 2015) (3) Two-stream fusion (C. Feichtenhofer et al, 2016.) (4) C3D (D. Tran et al, 2015) (5) Two-stream+I3D (J. Carreira and A. Zisserman, 2017) (6) TSN(L. Wang et al, 2016).

Overall, according Table1, compared with the Two-stream I3D and its pre-trained model, our 3 channels sub-model outperforms 0.5% and 0.7%, respectively.

In UTAR, Table 2 demonstrates that our architecture improves the untrimmed videos recognition about 3.5% on TUMOS14 thanks to the transfer learning, that combine the explicit knowledge from trimmed videos.

**Table 4** TAR accuracy that is the average accuracy over 3 splits of UCF101.

| Method | UCF101 |
|---|---|
| Two-stream CNN[11] | 88.0 |
| Two-stream+LSTM[6] | 88.6 |
| Two-stream Fusion[4] | 92.5 |
| C3D 3nets [14] | 90.4 |
| Two-stream+I3D[1] | 93.4 |
| Two-stream+I3D(kinetics pre-trained)[1] | 98.0 |
| TSN[18] | 94.0 |
| Two-stream+ multi-channels self-attention | 93.9 |
| Two-stream(kinetics pre-trained) + multi-channels self-attention | 98.7 |

**Table 5** UTAR accuracy on THUMOS14.

| Method | THUMOS14 |
|---|---|
| IDT+FV[16] | 63.1 |
| Two-stream[39] | 66.1 |
| Objects+Motion[7] | 66.8 |
| Depth2Action[21] | - |
| C3D[14] | - |
| TSN 3seg[18] | 78.5 |
| UntrimmedNet[17] | 82.2 |
| OUR | 89.7 |

## 5 Conclusion

In this paper, based on a pre-training multi-channels self-attention model, We propose a composite two-stream framework. The novel framework includes a classifier stream and a composite instance stream. The graph network model is adopted in each of the two streams, the performance of feature extraction is significantly improved, and reasoning ability of the framework becomes better. In the composite instance stream, a multi-channel self-attention models are constructed to weight each frame in the video and give more attention to the key frames. Each self-attention models channel outputs a set of attention weights to focus on a particular aspect of the video. The multi-channels self-attention models can evaluate key-frames from multiple aspects, and the output sets of attention weight vectors form an attention matrix. This model can implement action recognition under zero-shot conditions, and has good recognition performance for untrimmed video data. Experimental results on relevant data sets confirm the validity of our model.

## 6 Acknowledgments

We are very grateful to DeepBlue Technology (Shanghai) Co., Ltd. and DeepBlue Academy of Sciences for their support. Thanks to the support of the Equipment pre-research project (No. 31511060502). Thanks to Dr. Dongdong Zhang of the DeepBlue Academy of Sciences.